\documentclass{article}

\usepackage{microtype}
\usepackage{graphicx}
\usepackage{subfigure}
\usepackage{booktabs} 
\providecommand{\Description}[1]{}

\usepackage{hyperref}

\providecommand{\Require}{\REQUIRE}
\providecommand{\Ensure}{\ENSURE}
\providecommand{\State}{\STATE}
\providecommand{\Comment}[1]{\COMMENT{#1}}
\providecommand{\If}[1]{\IF{#1}}

\providecommand{\EndIf}{\ENDIF}
\providecommand{\For}[1]{\FOR{#1}}
\providecommand{\ForAll}[1]{\FORALL{#1}}
\providecommand{\EndFor}{\ENDFOR}
\providecommand{\While}[1]{\WHILE{#1}}
\providecommand{\EndWhile}{\ENDWHILE}
\providecommand{\Function}[2]{\FUNCTION{#1(#2)}}
\providecommand{\EndFunction}{\ENDFUNCTION}
\providecommand{\Return}{\textbf{return}\ }
\providecommand{\Call}[2]{\textsc{#1}(#2)}

\usepackage[accepted]{icml2025}

\usepackage{amsmath}
\usepackage{amssymb}
\usepackage{mathtools}
\usepackage{amsthm}

\usepackage[capitalize,noabbrev]{cleveref}

\theoremstyle{plain}

\theoremstyle{definition}

\theoremstyle{remark}

\usepackage[textsize=tiny]{todonotes}

\icmltitlerunning{The Shape of Reasoning: Topological Analysis of Reasoning Traces in Large Language Models}

\begin{document}

\twocolumn[
\icmltitle{The Shape of Reasoning: Topological Analysis of \\ Reasoning Traces in Large Language Models}



\icmlsetsymbol{equal}{*}

\begin{icmlauthorlist}
\icmlauthor{Xue Wen Tan}{comp,sch}
\icmlauthor{Galen Lee}{equal,comp,yyy,zzz}
\icmlauthor{Nathaniel Tan}{equal,comp,yyy,zzz}
\icmlauthor{Stanley Kok}{sch}
\end{icmlauthorlist}

\icmlaffiliation{yyy}{University of Cambridge, Department of Engineering, England}
\icmlaffiliation{comp}{Infocomm Media Development Authority, BizTech Group, Singapore}
\icmlaffiliation{sch}{National University of Singapore, School of Computing, Singapore}
\icmlaffiliation{zzz}{Work Done During Internship}

\icmlcorrespondingauthor{Xue Wen Tan}{xuewen@u.nus.edu}

\icmlkeywords{LLM Reasoning, Reasoning Evaluation, Automated Evaluation, Topological Data Analysis}

\vskip 0.3in
]



\printAffiliationsAndNotice{\icmlEqualContribution} 

\begin{abstract}
Evaluating the quality of reasoning traces from large language models remains understudied, labor-intensive, and unreliable: current practice relies on expert rubrics, manual annotation, and slow pairwise judgments. Automated efforts are dominated by graph-based proxies that quantify structural connectivity but do not clarify what constitutes high-quality reasoning; such abstractions can be overly simplistic for inherently complex processes. We introduce a topological data analysis (TDA)–based evaluation framework that captures the geometry of reasoning traces and enables label-efficient, automated assessment. In our empirical study, topological features yield substantially higher predictive power for assessing reasoning quality than standard graph metrics, suggesting that effective reasoning is better captured by higher-dimensional geometric structures rather than purely relational graphs. We further show that a compact, stable set of topological features reliably indicates trace quality, offering a practical signal for future reinforcement learning algorithms.
\end{abstract}
\section{Introduction}

Large language models (LLMs) have demonstrated remarkable reasoning capabilities across diverse domains and tasks, including commonsense reasoning, mathematical reasoning, logical reasoning, causal reasoning, visual reasoning, audio reasoning, multimodal reasoning, embodied reasoning, and defeasible reasoning \cite{sun2023survey}. However, despite their impressive performance, the internal mechanisms underlying their reasoning processes remain poorly understood. Intermediate reasoning steps are often overlooked and treated merely as a means to the final answer rather than as objects of study in their own right. For instance, \citet{wangself} marginalizes diverse reasoning paths by aggregating multiple solutions and selecting the most consistent final answer, thereby discarding valuable information about the reasoning process itself. Moreover, there are growing concerns about the faithfulness of intermediate reasoning steps to the final answer \cite{agarwal2024faithfulness}. LLMs may arrive at correct answers through flawed or spurious reasoning processes, raising questions about whether the generated reasoning genuinely reflects the model's decision-making or simply provides post-hoc justification for memorized answers.

Evaluating the quality of reasoning processes presents significant challenges. First, there is a scarcity of datasets that provide step-by-step solutions; most datasets offer only final answers, forcing researchers to use answer correctness as an imperfect proxy for reasoning quality. Second, assessing reasoning quality is inherently subjective, and developing reliable quantitative metrics remains a complex and unresolved problem.

To address these challenges, we introduce a novel evaluation framework grounded in a carefully curated dataset derived from the American Invitational Mathematics Examination (AIME) available on the Art of Problem Solving portal \cite{AoPS_AIME}. This dataset includes detailed solution steps contributed by mathematicians in the community. Importantly, we incorporate multiple solution paths for each problem, recognizing that mathematical problems can be solved through diverse reasoning strategies. However, we acknowledge that even expert-written solutions represent only the externalized steps of reasoning, omitting the intermediate cognitive processes required to progress from one step to the next. To capture a fuller picture of the reasoning process, we adapt the Smith-Waterman algorithm from biological sequence alignment \cite{smith1981identification} to align LLM-generated reasoning traces with reference solutions, addressing the first challenge of dataset scarcity.

To address the second challenge of subjective assessment, we employ topological data analysis (TDA) to quantify reasoning quality objectively. We hypothesize that high-quality reasoning traces exhibit distinctive structural characteristics that persist across diverse solution strategies. Topology provides a principled framework for capturing these invariant properties: just as a coffee mug and a donut are topologically equivalent (homeomorphic) despite their apparent differences \cite{gamelin1999introduction}, diverse valid reasoning paths may share fundamental structural similarities that distinguish them from flawed reasoning. TDA enables us to identify and measure these invariant features, providing objective metrics for reasoning quality that are robust to surface-level variations in problem-solving approaches. 

Overall, our contributions include:
\begin{itemize}
\item A novel application of the Smith-Waterman algorithm to align and compare LLM reasoning traces with expert solutions, enabling systematic evaluation on curated AIME problems with multiple solution paths.
\item A topological data analysis framework for objectively quantifying reasoning quality.
\item Empirical evidence that topological features provide discriminative signals for distinguishing high-quality from flawed reasoning traces.
\end{itemize}
\section{Related Work}

\subsection{Reasoning Traces Analysis}
The analysis of reasoning traces in large language models (LLMs) has become a central research direction following the introduction of chain-of-thought (CoT) prompting \cite{wei2022chain}, which improved problem-solving performance by encouraging models to produce intermediate reasoning steps. However, Self-Consistency prompting \cite{wangself} and related methods treat reasoning traces as instrumental by aggregating multiple reasoning paths to obtain the most frequent final answer while discarding the traces themselves. Recent work has shifted attention to reasoning faithfulness: whether a model’s explicit reasoning aligns with its underlying decision process. \citet{agarwal2024faithfulness} demonstrated that LLMs often produce plausible but unfaithful explanations that do not reflect actual reasoning, and \citet{nguyen2024direct} found that models can achieve correct answers through logically flawed or spurious reasoning. These findings emphasize the need for metrics that evaluate reasoning process quality rather than final-answer correctness.

To quantify reasoning structure, \citet{xiong2025mapping} introduced a reasoning-graph framework that maps LLM-generated reasoning traces into directed graphs, enabling analysis of properties such as branching and convergence. Their findings suggest that effective reasoning depends on a balanced pattern of exploration and convergence rather than on trace length alone, aligning with the observations of \citet{su2025between} that overly long reasoning often reduces accuracy. Complementary work has explored automated reasoning evaluation: \citet{tonunderstanding} proposed an information-theoretic approach measuring the contribution of each reasoning step, while \citet{nguyen2024direct} grounded CoT reasoning in knowledge graphs to test factual and logical validity. Together, these works advance the field toward systematic evaluation of reasoning traces.

Efforts to improve reasoning quality have paralleled those to analyze it. Reinforcement learning methods, such as GraphPRM \cite{peng2025rewarding}, directly optimize reasoning steps rather than outcomes, using process-level rewards to align model behavior with logical norms. Structured prompting techniques such as Tree-of-Thoughts \cite{yao2023tree} and Graph-of-Thoughts \cite{besta2024graph} further demonstrate that the organization of reasoning whether tree, graph, or chain-based may critically influences problem-solving success. Collectively, these studies motivate the search for more robust and interpretable reasoning metrics, which our work seeks to advance through a geometric–topological formulation.

\subsection{Topological Data Analysis Applied to LLMs}

Topological Data Analysis (TDA) offers a mathematical lens for understanding the shape of data by identifying invariant geometric structures such as connected components and holes. Recent work has applied TDA to study neural representations in large models. \citet{gardinazzipersistent} introduced zigzag persistent homology to analyze the evolution of topological features across transformer layers, uncovering distinct representational phases and enabling topology-guided layer pruning. \citet{ruppik2024local} examined the local topology of contextual embedding spaces, showing that persistence-based neighborhood descriptors improve linguistic disambiguation. \citet{balderas2025green} used persistent homology to measure neuron importance in BERT, achieving substantial compression gains while maintaining performance and providing explainability through persistence diagrams. Collectively, these studies suggest that topological descriptors can capture structural properties of high-dimensional representations that traditional metrics overlook. Our work extends this line of inquiry by applying TDA to reasoning traces, treating them as geometric objects whose topological structure encodes reasoning quality. By quantifying persistent topological features across reasoning paths, we provide an interpretable and label-efficient approach to automated reasoning evaluation.

\section{Methodology}
\label{sect:Methodology}

We evaluate the \emph{shape} of LLM reasoning traces in four stages: (1) generate traces for AIME problems, (2) align model and expert steps in embedding space, (3) extract topological features from those embeddings, and (4) compute graph-theoretic baselines on the same data.

\subsection{Stage 1: Generating Reasoning Traces}

We parse AIME JSON into tuples $(x_i,s_i)$ where $x_i$ is the problem statement and $s_i$ is the reference solution text from AoPS. We then generate model traces $r_i$ with an answer-blind prompt through a local Ollama endpoint and store outputs in JSONL format (prompt details in Appendix \ref{app:prompt}).

\subsection{Stage 2: Aligning Model Steps to Gold}

We segment both $r_i$ and $s_i$ into step lists $R_i=(r_{i,1},\dots,r_{i,m})$ and $S_i=(s_{i,1},\dots,s_{i,n})$ using a rule-based segmenter. Each step is embedded with \texttt{all-mpnet-base-v2} to produce
\[
X^{(r)}_i=\begin{bmatrix} \mathbf{x}^{(r)}_{i,1} \\ \vdots \\ \mathbf{x}^{(r)}_{i,m}\end{bmatrix}\in\mathbb{R}^{m\times d},
\qquad
X^{(s)}_i=\begin{bmatrix} \mathbf{x}^{(s)}_{i,1} \\ \vdots \\ \mathbf{x}^{(s)}_{i,n}\end{bmatrix}\in\mathbb{R}^{n\times d}.
\]

We compute local Smith--Waterman alignment in embedding space. With match score $s_{uv}$ and gap penalty $\gamma>0$, the dynamic-programming recurrence is
\begin{equation}
\begin{aligned}
H_{u,v}&=\max\Bigl\{\,0,\;H_{u-1,v-1}+s_{uv},\\
&\qquad H_{u-1,v}-\gamma,\;H_{u,v-1}-\gamma\Bigr\},\\
H_{0,\cdot}&=H_{\cdot,0}=0.
\end{aligned}
\end{equation}
Backtracking from $\arg\max H_{u,v}$ returns aligned index pairs $\mathcal{A}_i$ plus two summary metrics: mean alignment score and gold-step coverage. A concrete alignment example is shown in Appendix \ref{app:swad}.

\subsection{Stage 3: Topological Feature Extraction}

Given embedded steps $X=\{\mathbf{x}_1,\ldots,\mathbf{x}_\ell\}\subset\mathbb{R}^d$, we define cosine distance as
\begin{equation}
    \mathrm{dist}(\mathbf{x}_p,\mathbf{x}_q)=1-\dfrac{\langle \mathbf{x}_p,\mathbf{x}_q\rangle}{\|\mathbf{x}_p\|\,\|\mathbf{x}_q\|}
\end{equation}

We build Vietoris--Rips filtrations and compute persistence diagrams
\[
\mathcal{D}_k=\{(b_j^{(k)},d_j^{(k)})\}_{j=1}^{N_k},\quad 0\le k\le \texttt{maxdim}.
\]
We focus on $k\in\{0,1\}$ (connected components and 1-cycles).

From $\{\mathcal{D}_k\}$ we extract three feature families: (i) VR summary statistics, (ii) Betti-curve descriptors, and (iii) persistence landscape descriptors (full definitions in Appendix \ref{app:tdafeatures}).

\subsection{Stage 4: Graph Baselines}

For comparability with prior work, we also compute graph features on the same embeddings following \citet{minegishi2025topology}. We construct per-trace graphs and extract \texttt{has\_loop}, \texttt{loop\_count}, diameter, average clustering $\overline{C}$, average shortest path $\overline{L}$, and small-world index. Feature definitions are listed in Appendix \ref{app:graphfeatures}.
\section{Experimental Results}
\label{sec:results}

\subsection{Predictive Power: Topology vs. Graph}

\begin{table*}[htbp]
\centering
\caption{\textbf{Feature Sets vs. Smith--Waterman Alignment Score} (AIME 2020--2025, combined).
Each model has 180 observations. Metrics are $R^2$, adjusted $R^2$, and (\%) gains over the TDA-only baseline ($\Delta$ columns); higher is better.}
\label{tab:sw-aime}
\footnotesize
\setlength{\tabcolsep}{4pt}
\resizebox{\textwidth}{!}{%
\begin{tabular}{l cc cc cccc}
\toprule
Model
  & \multicolumn{2}{c}{Graph}
  & \multicolumn{2}{c}{TDA}
  & \multicolumn{4}{c}{Graph + TDA} \\
\cmidrule(lr){2-3}\cmidrule(lr){4-5}\cmidrule(lr){6-9}
 & $R^2$ & Adj.\ $R^2$ & $R^2$ & Adj.\ $R^2$ & $R^2$ & $\Delta R^2$ vs TDA (\%) & Adj.\ $R^2$ & $\Delta$ Adj.\ $R^2$ vs TDA (\%) \\
\midrule
Qwen3-8B & 0.054 & 0.021 & \textbf{0.273} & \textbf{0.154} & 0.312 & 14.3 & 0.167 & 8.4 \\
Qwen3-32B & 0.088 & \textbf{0.056} & \textbf{0.181} & 0.048 & 0.233 & 28.7 & 0.073 & 52.1 \\
Qwen3-235B & 0.024 & -0.004 & \textbf{0.163} & \textbf{0.027} & 0.167 & 2.5 & -0.001 & -103.7 \\
DeepSeek-r1-7B$^{\dagger}$ & 0.047 & 0.014 & \textbf{0.210} & \textbf{0.082} & 0.226 & 7.6 & 0.063 & -23.2 \\
DeepSeek-r1-32B & 0.057 & 0.024 & \textbf{0.190} & \textbf{0.059} & 0.226 & 18.9 & 0.064 & 8.5 \\
DeepSeek-r1-70B & 0.058 & 0.025 & \textbf{0.249} & \textbf{0.127} & 0.300 & 20.5 & 0.153 & 20.5 \\
GPT-OSS-20B & 0.081 & 0.049 & \textbf{0.296} & \textbf{0.182} & 0.327 & 10.5 & 0.186 & 2.2 \\
GPT-OSS-120B & 0.101 & 0.070 & \textbf{0.327} & \textbf{0.218} & 0.368 & 12.5 & 0.236 & 8.3 \\
\bottomrule
\multicolumn{9}{l}{\footnotesize $^{\dagger}$~DeepSeek-r1-7B was chosen instead of 8B to maintain consistency among models, since 8B is derived from Qwen3 while the others originate from Qwen2.5.}
\end{tabular}
}
\end{table*}

We fit OLS regressions to predict Smith--Waterman alignment scores using three feature sets: Graph only, TDA only, and Graph+TDA. Across eight model settings, TDA consistently outperforms Graph as a standalone predictor (mean $R^2$: 0.236 vs. 0.064; mean adjusted $R^2$: 0.112 vs. 0.032), and TDA has higher adjusted $R^2$ in 7/8 settings.

Adding Graph to TDA gives modest gains in raw fit (mean $\Delta R^2=+14.4\%$) but inconsistent complexity-adjusted gains (mean $\Delta$ adjusted $R^2=-3.4\%$). In two settings (Qwen3-235B and DeepSeek-r1-7B), adjusted fit decreases after adding Graph features. Overall, topology is the stronger and more stable signal of reasoning quality.

\subsection{Significant Topological Features}

Because the 28 raw TDA features are strongly collinear (Appendix Table~\ref{tab:vif}), we cluster them by correlation structure and regress alignment on cluster representatives. The full clustering pipeline, silhouette analysis, cluster-feature mapping, and complete regression table can be found in Appendix~\ref{app:clustered_tda}.

Across all models, four cluster-level effects are consistently informative:
\begin{itemize}
    \item \textbf{Cluster 2 (H0 betti\_spread):} positive association with alignment.
    \item \textbf{Cluster 3 (H0 betti\_width):} negative association with alignment.
    \item \textbf{Cluster 12 (H1 betti\_width):} positive association with alignment.
    \item \textbf{Cluster 16 (H1 max\_birth/max\_death):} weak negative association.
\end{itemize}

In practical terms, higher-quality traces tend to maintain a coherent main line, include short and varied local checks, and avoid late, large-scale detours.

\section{TDA Features Relations to Graph Features}

Graph statistics such as clustering, path length, diameter, loop count, and the small-world index are intuitive, but they compress geometry and thus lack granularity. By contrast, TDA features capture these nuances: $H_0$ descriptors capture how tightly points clump and merge (component geometry), while $H_1$ descriptors capture the presence and persistence of detours/cycles. Regressing graph features on TDA (Table~\ref{tab:tdavsgraph}) lets us translate between the two. The $R^2$ row shows that TDA accounts for a sizeable share of variance in clustering, path length, diameter, and the small-world index ($\approx .35$-$.38$), but much less in loop incidence ($\approx .07$), indicating that global cohesion/efficiency is strongly topological, whereas loop multiplicity is more idiosyncratic. Below we discuss each graph feature, focusing on the most prominent effects ($p<.01$ and comparatively larger magnitude) and interpreting how the implicated TDA signals generate the corresponding graph patterns. 

\paragraph{Graph Average clustering.}
Clustering rises when $H_0$ components persist longer on average (\texttt{H0 mean life} $+$), because durable local pockets (a tight local cluster of nearby points = a group of points whose pairwise distances are small—so in VR they connect at a small radius and stay separate from other points until a larger radius) make it easy to close \emph{triangles} (three nodes all mutually linked: $i$–$j$, $i$–$k$, and $j$–$k$). By contrast, clustering decreases when the center of $H_0$ activity shifts to larger filtration scales (\texttt{H0 betti centroid} $-$): if most merging is deferred until later, fine-scale neighbor–neighbor links are sparser and local closure is harder to realize.

\paragraph{Graph Average path length.}
Distances shrink when components live longer (\texttt{H0 mean life} $-$) and when merging is distributed across scales (\texttt{H0 betti spread} $-$). A larger $H_0$ spread means connectivity appears at both fine and intermediate scales; in graph terms this corresponds to having not only very short transitions within neighborhoods but also intermediate “bridges” between neighboring ones, further shortening routes. By contrast, distances grow when most $H_0$ activity occurs at larger scales (\texttt{H0 betti centroid} $+$), because early fine-scale links are missing, and when $H_0$ lifetimes are uneven (\texttt{H0 entropy} $+$), because early-merged regions sit next to late-merged ones, creating bottlenecks.

\paragraph{Graph Diameter.}
The same drivers that shorten (or lengthen) typical paths also contract (or inflate) extremal distances, with amplified effects. Diameter inflates when $H_0$ activity is centered late (\texttt{H0 betti centroid} $+$), because the farthest nodes lack early bridges. It deflates when pockets persist and stitching is multi-scale (\texttt{H0 mean life} $-$; \texttt{H0 betti spread} $-$), which caps the longest geodesics by adding early local and mid-range links.

\paragraph{Graph Loop count.}
Loop multiplicity is the least predictable target (low $R^2$), but two signals are robust. A stronger $H_1$ cycle signal increases loop count (\texttt{H1 landscape mean} $+$): more/persistent 1-cycles create structural detours that let the trajectory leave a region and then re-enter it, producing extra revisits of the first repeated node. In contrast, cohesive $H_0$ structure suppresses loops (\texttt{H0 mean life} $-$): when local components persist, the path tends to progress through many nearby–but–new nodes within a pocket before moving on, lowering the chance of returning multiple times to exactly the same node.

\paragraph{Graph Small-world index.}
Small-worldness rises with durable local structure (\texttt{H0 mean life} $+$) and falls when connectivity shifts to coarser scales (\texttt{H0 betti centroid} $-$). Intuitively, longer-lived pockets lift clustering $C$ early and, once a few cross-pocket links appear, restrain path length $L$ relative to random baselines, increasing $\sigma=\frac{C/C_{\mathrm{rand}}}{L/L_{\mathrm{rand}}}$. Conversely, a later $H_0$ centroid delays both local closure and shortcuts, depressing $\sigma$.

\paragraph{Takeaway.}
The \emph{centroid} says \textit{when} most merging happens along the scale (earlier vs.\ later); \emph{mean life} says \textit{how cohesive} local pockets are (longer lifetimes = tighter pockets); \emph{spread} says \textit{over how wide a range of scales} merging occurs \emph{(narrow scale = merging concentrated in a tight band of filtration radii; wide scale = merging spread across a broad band of radii, so both very short and intermediate links appear)}; and \emph{entropy} says \textit{how uniform} those lifetimes are (even vs.\ uneven). Earlier merging and longer lifetimes raise clustering and shorten paths; wide-scale merging further shortens paths and limits diameter; later merging or uneven lifetimes lengthen paths and lower the small-world index. $H_1$ then controls cycle-richness: stronger $H_1$ increases loopiness, while cohesive $H_0$ tends to suppress it. This is why clustering, path length, diameter, and small-worldness closely track $H_0$, whereas raw loop counts depend more on small connection patterns among a few nodes.

\section{Limitations and Future Work}

Despite demonstrating that topological summaries can capture structural properties of reasoning traces and correlate more strongly with alignment quality than classical graph metrics, our work has several limitations that point to directions for future research.

\paragraph{Dataset scope and generality.} Our experiments use the American Invitational Mathematics Examination (AIME) dataset because, to the best of our knowledge, it is the only publicly available corpus with step–by–step solution traces for non‑trivial problems.  While olympiad‑level math offers rich reasoning chains, this focus restricts the diversity of reasoning styles and problem domains considered. Surveys on step‑by‑step reasoning evaluation note that existing resources tend to be either overly simple or confined to specialised domains \cite{lee2025evaluating}. Relying on a single dataset therefore limits the generality of our findings. A clear avenue for future work is to curate or annotate additional datasets with explicit reasoning traces across domains such as commonsense reasoning, science, programming and real‑world problem solving. Even small human‑annotated corpora in other domains would allow us to test whether the topological patterns identified here persist beyond competitive mathematics.

\paragraph{Topological features interpretation.} Our features summarize the geometry of step embeddings, not the symbolic structure of reasoning trees. Holes in $H_1$ and mergers in $H_0$ arise from how steps are placed in the embedding space and from the distance used to build the Vietoris–Rips complex; they need not correspond to literal detours or merges in a human-readable proof. In our pipeline, both alignment and topology operate on sentence embeddings with cosine distance. Changing the embedder, the segmentation, or the metric can create or remove cycles and shift lifetimes without altering the underlying textual logic. Consequently, interpretations such as ``$H_1$ captures detours'' or ``$H_0$ captures clustering of ideas'' are, at best, geometric proxies for reasoning structure rather than direct evidence of a particular tree or graph motif. We therefore caution against reading persistence diagrams as faithful maps of the latent reasoning program and instead view them as embedding-dependent signals that correlate with alignment quality. For future work, we aim to ground topological events in interpretable operations, such as opening a branch, running a short check, and rejoining, while remaining graph-free, since explicit trace graphs are rarely available.

\section*{Acknowledgements}
We would like to thank Infocomm Media Development Authority (IMDA) for providing us with the opportunity and funding to work on this research project. Any opinions, findings, and conclusions or recommendations expressed in this material are those of the author(s) and do not reflect the views of the Infocomm Media Development Authority, Singapore.

\bibliography{main}
\bibliographystyle{icml2025}

\newpage
\appendix
\onecolumn
\section{Correlation Tables for TDA Features}
\begin{figure}[H]
  \centering
  \includegraphics[width=\linewidth]{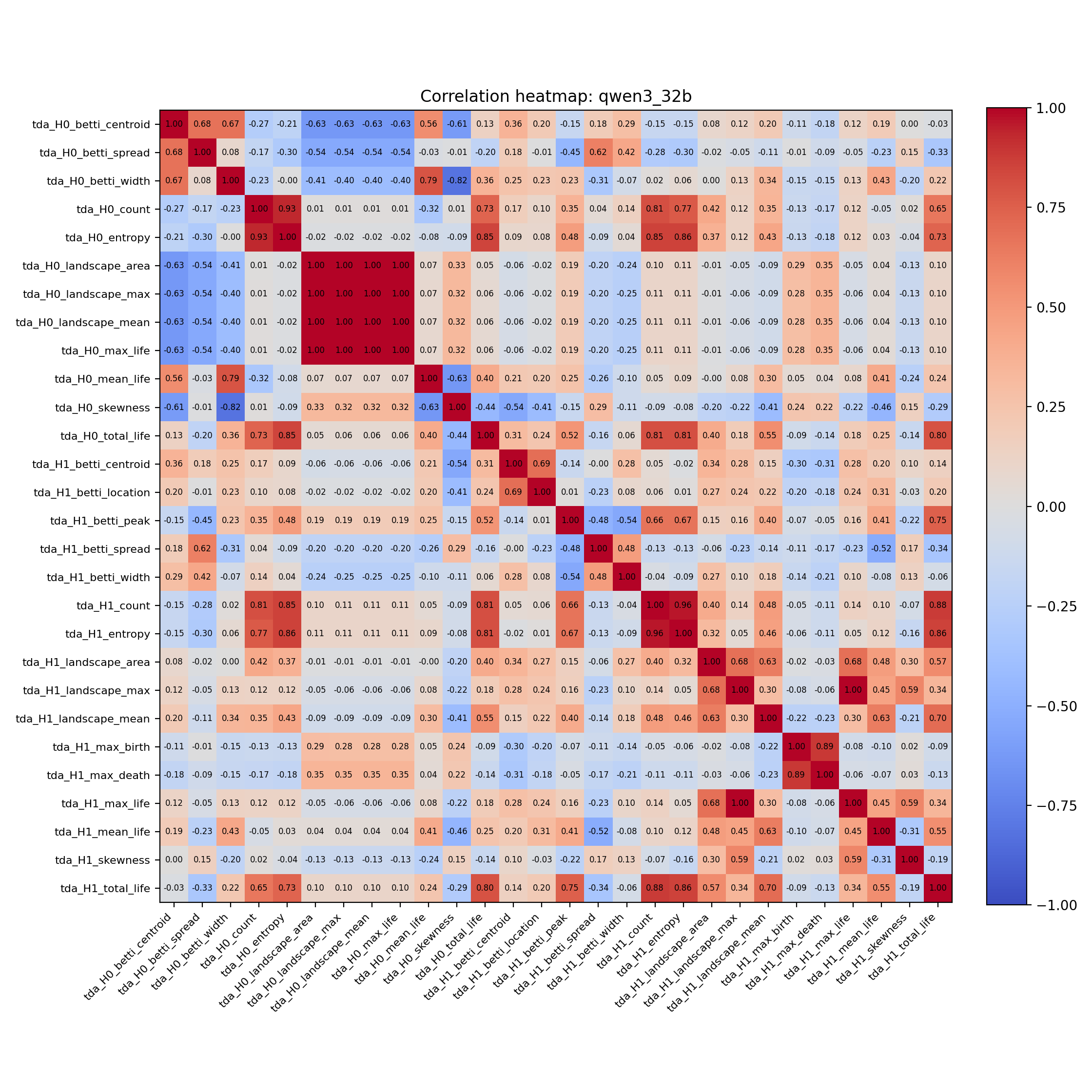}
  \caption{Correlation Heatmap for Qwen3-32B.}
  \Description{}
\end{figure}

\begin{figure}[H]
  \centering
  \includegraphics[width=\linewidth]{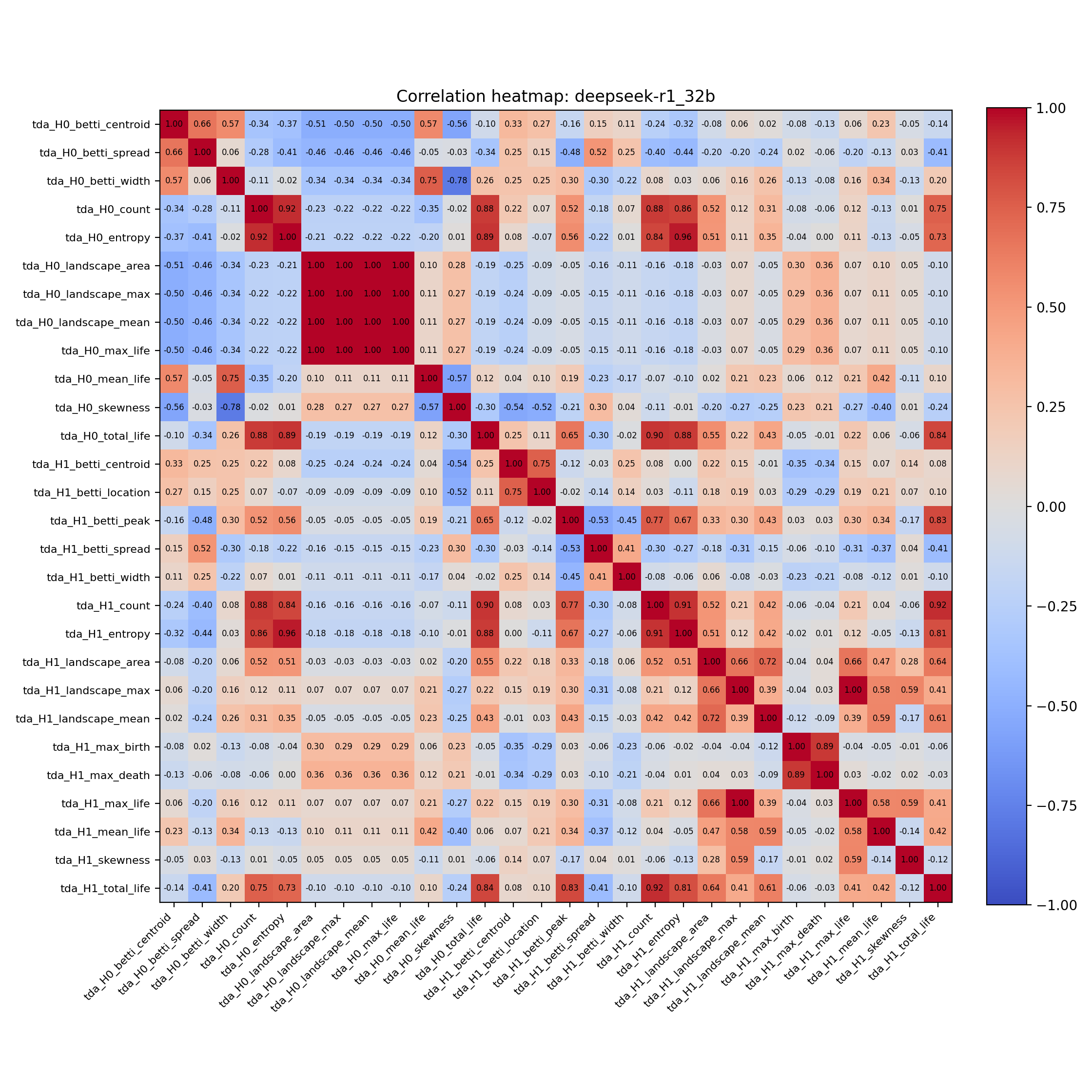}
  \caption{Correlation Heatmap for Deepseek-r1-32B.}
  \Description{}
\end{figure}

\begin{figure}[H]
  \centering
  \includegraphics[width=\linewidth]{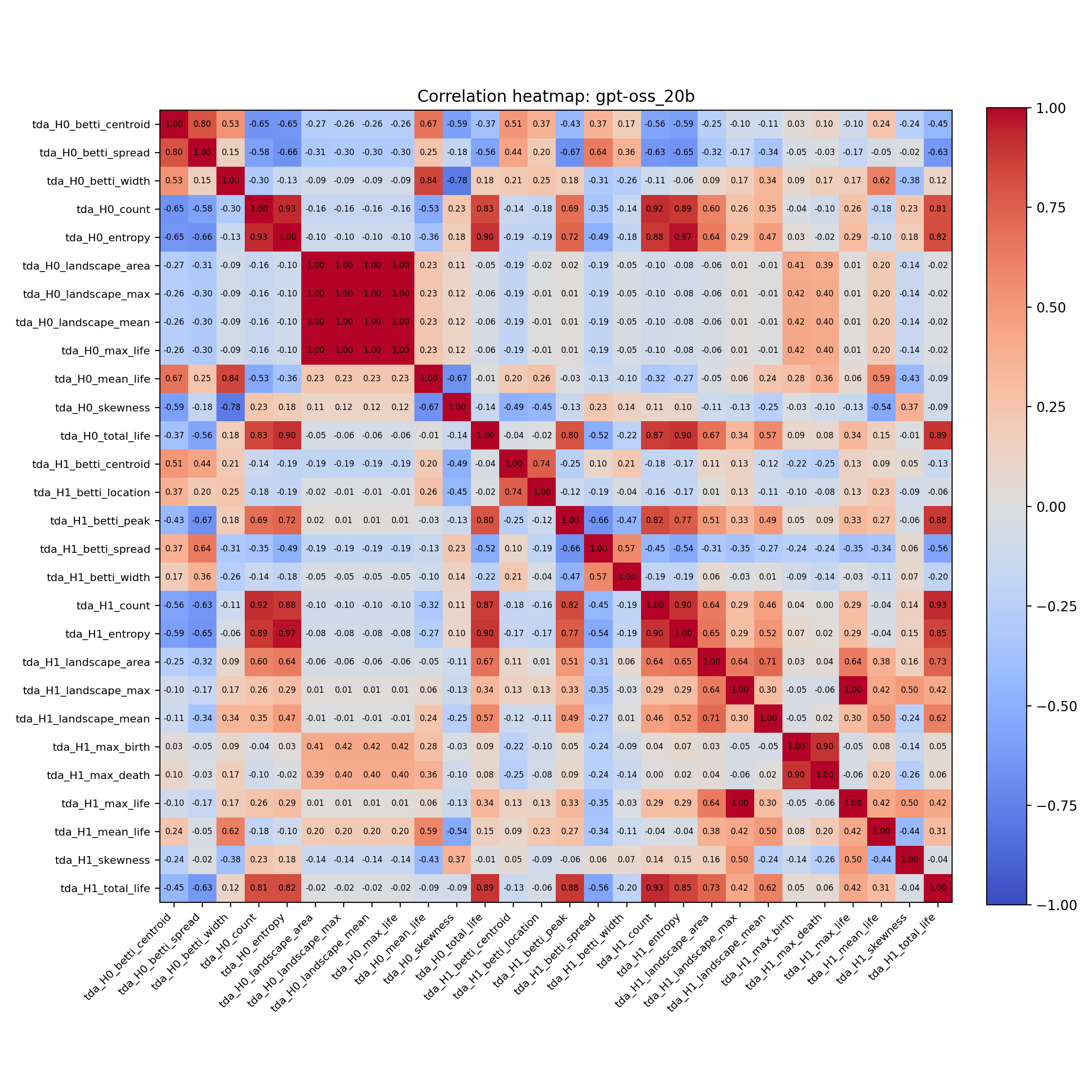}
  \caption{Correlation Heatmap for GPT-OSS-20B.}
  \Description{}
\end{figure}
\section{Topological Profiles of Various Small-Mid Sized Models}
\label{app:topoprof}

\begin{figure}[H]
  \centering
  \includegraphics[width=0.6\linewidth]{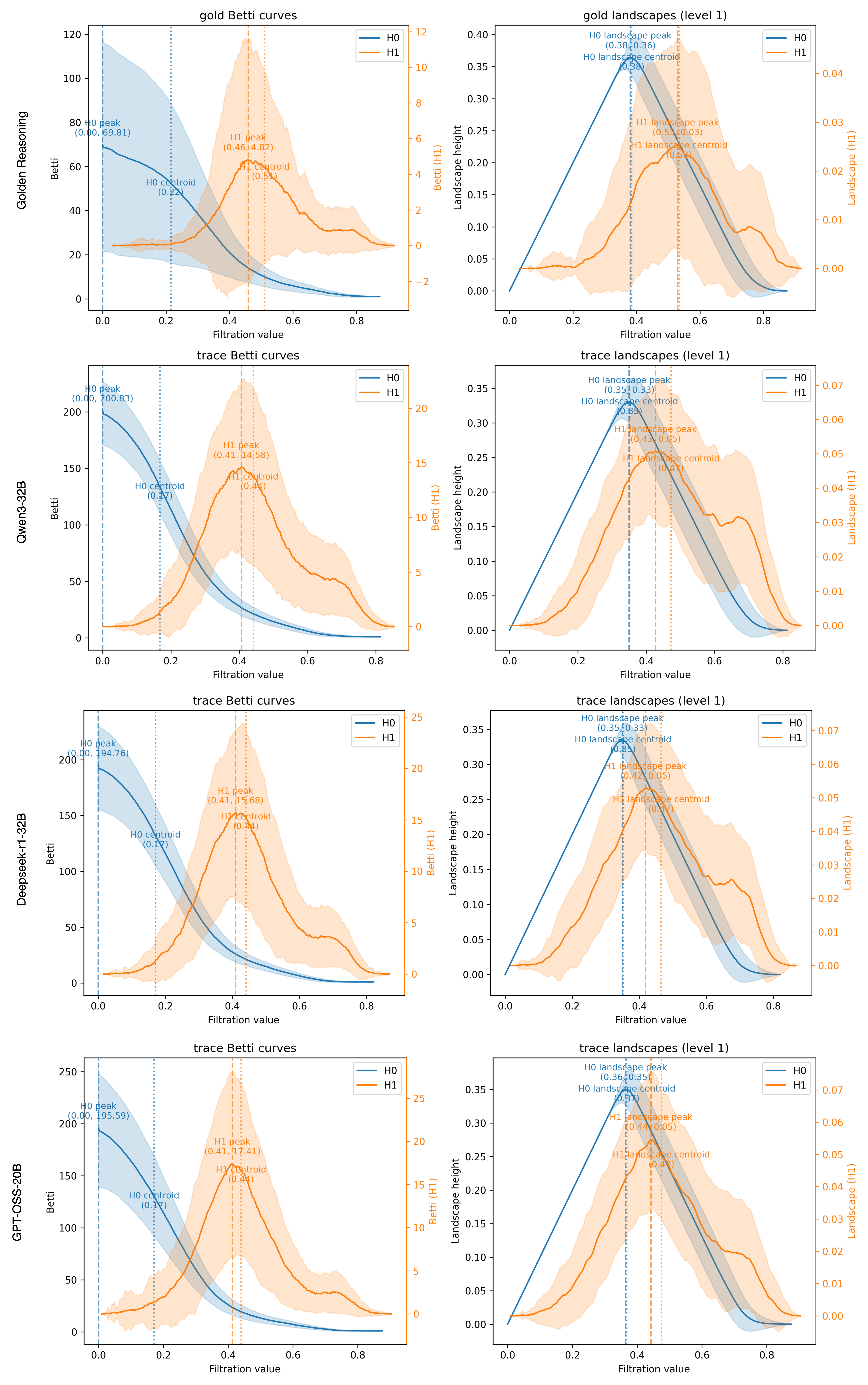}
  \caption{Visualization of Betti Curves and Persistence Landscapes for Qwen3-32B, Deepseek-r1-32B, and GPT-OSS-20B models. The plots show the mean and standard deviation of topological features (Betti numbers and persistence landscapes) computed from reasoning traces on AIME datasets (2020-2025). Each row corresponds to a model, with the left panels displaying Betti curves (for H0 and H1) and the right panels showing first-level persistence landscapes. Peaks and centroids for H0 and H1 are annotated to highlight key topological characteristics across the reasoning processes.}
  \Description{}
\end{figure}

\onecolumn
\section{Topological Features Descriptions}
\label{app:tdafeatures}

\begin{table}[H]
\small
\centering
\caption{TDA feature dictionary (use $k=0,1,\ldots$; e.g., \texttt{H0\_*}, \texttt{H1\_*}).Removed H0\_betti\_peak (same as H0\_count), H0\_betti\_location \& H0\_max\_birth (both are constant), H0\_max\_death (equal to H0\_max\_life)}
\renewcommand{\arraystretch}{1.15}
\begin{tabular}{p{0.3\linewidth} p{0.6\linewidth}}
\hline
\textbf{TDA feature} & \textbf{Description} \\
\hline
\multicolumn{2}{l}{\textit{Vietoris--Rips diagram features}} \\
\texttt{H\{k\}\_count} & Number of intervals (points) in the persistence diagram for homology $H_k$. \\
\texttt{H\{k\}\_total\_life} & Sum of lifetimes $\sum(\text{death}-\text{birth})$ across intervals (intervals with $\infty$ death contribute $0$). \\
\texttt{H\{k\}\_max\_life} & Maximum lifetime among intervals. \\
\texttt{H\{k\}\_mean\_life} & Mean lifetime across intervals (computed as total\_life divided by number of intervals). \\
\texttt{H\{k\}\_entropy} & Shannon entropy of the normalized positive-lifetime distribution. \\
\texttt{H\{k\}\_skewness} & Skewness of interval lifetimes. \\
\texttt{H\{k\}\_max\_birth} & Largest \emph{birth} filtration value among the $H_k$ intervals (i.e., the latest scale at which any $H_k$ feature first appears). For Vietoris--Rips on a point cloud, $H_0$ births are 0, so \texttt{H0\_max\_birth}$=0$. \\
\texttt{H\{k\}\_max\_death} & Largest \emph{finite death} filtration value among the $H_k$ intervals (i.e., the latest scale at which any $H_k$ feature is killed), ignoring intervals with death $=\infty$ that don’t die within the chosen VR threshold. For $H_0$ in VR, this equals \texttt{H0\_max\_life} because births are 0. \\

\addlinespace
\multicolumn{2}{l}{\textit{Betti-curve features}} \\
\texttt{H\{k\}\_betti\_peak} & Maximum Betti count (peak height of the Betti curve). \\
\texttt{H\{k\}\_betti\_location} & Normalized location of the peak in $[0,1]$ over the curve’s time range. \\
\texttt{H\{k\}\_betti\_width} & Normalized full width at half maximum (FWHM) in $[0,1]$. \\
\texttt{H\{k\}\_betti\_centroid} & Normalized first moment (center of mass) of the Betti curve. \\
\texttt{H\{k\}\_betti\_spread} & Normalized standard deviation of the curve about its centroid. \\
\addlinespace
\multicolumn{2}{l}{\textit{Persistence-landscape features}} \\
\texttt{H\{k\}\_landscape\_mean} & Mean height of the level-0 persistence landscape for $H_k$. \\
\texttt{H\{k\}\_landscape\_max} & Maximum height of the level-0 persistence landscape. \\
\texttt{H\{k\}\_landscape\_area} & Area under the level-0 persistence-landscape curve (trapezoidal rule). \\
\hline
\end{tabular}
\end{table}

\section{Graph Features Descriptions}
\label{app:graphfeatures}

\begin{table}[H]
\small
\centering
\caption{Graph features following \citet{minegishi2025topology}.}
\renewcommand{\arraystretch}{1.15}
\begin{tabular}{p{0.3\linewidth} p{0.6\linewidth}}
\hline
\textbf{Graph feature} & \textbf{Description} \\
\hline
\texttt{has\_loop} & Indicator (0/1) of whether the path revisits any node (cycle detected by first repeated node).%
\footnotesize~[cycle detection logic] \normalsize \\
\texttt{loop\_count} & Number of revisits of the \emph{first} repeated node (occurrences minus one).%
\footnotesize~[computed at first repeat] \normalsize \\
\texttt{diameter} & Graph geodesic diameter: the maximum over nodes of the shortest-path distances (Dijkstra on the weighted, directed adjacency built from consecutive path steps). \\
\texttt{avg\_path\_length} & Mean of all finite shortest-path distances across nodes in that directed graph. \\
\texttt{avg\_clustering} & Average local clustering coefficient computed on the undirected projection: for nodes with degree $\ge 2$, the fraction of realized neighbor–neighbor edges, averaged over such nodes. \\
\texttt{small\_world\_index} & Small-world index $\sigma=\frac{C/C_{\mathrm{rand}}}{L/L_{\mathrm{rand}}}$ using $C_{\mathrm{rand}}=\tfrac{K}{N-1}$ and $L_{\mathrm{rand}}=\tfrac{\log N}{\log K}$ (when defined), where $N$ is node count and $K$ the mean degree. \\
\hline
\end{tabular}
\end{table}

\newpage
\section{System Prompt to Reasoning Models}
\label{app:prompt}

We employed Ollama (\url{https://ollama.com/}) to run the large language models on our local machine with Quadro RTX 8000. We have provided the following system prompt to generate the reasoning traces and output:

\begin{verbatim}
    SYS_PROMPT_WITHOUT_ANSWER = (
    "You are an expert competition mathematician.
    Solve the problem carefully and 
    present a clear, rigorous step-by-step solution. 
    Ensure each step is justified and consistent. 
    End with the final line formatted
    exactly as 'Final Answer: <number>'.")
\end{verbatim}

\section{Smith-Waterman Alignment Diagram}
\label{app:swad}

\begin{figure}[H]
  \centering
  \includegraphics[width=0.74\linewidth]{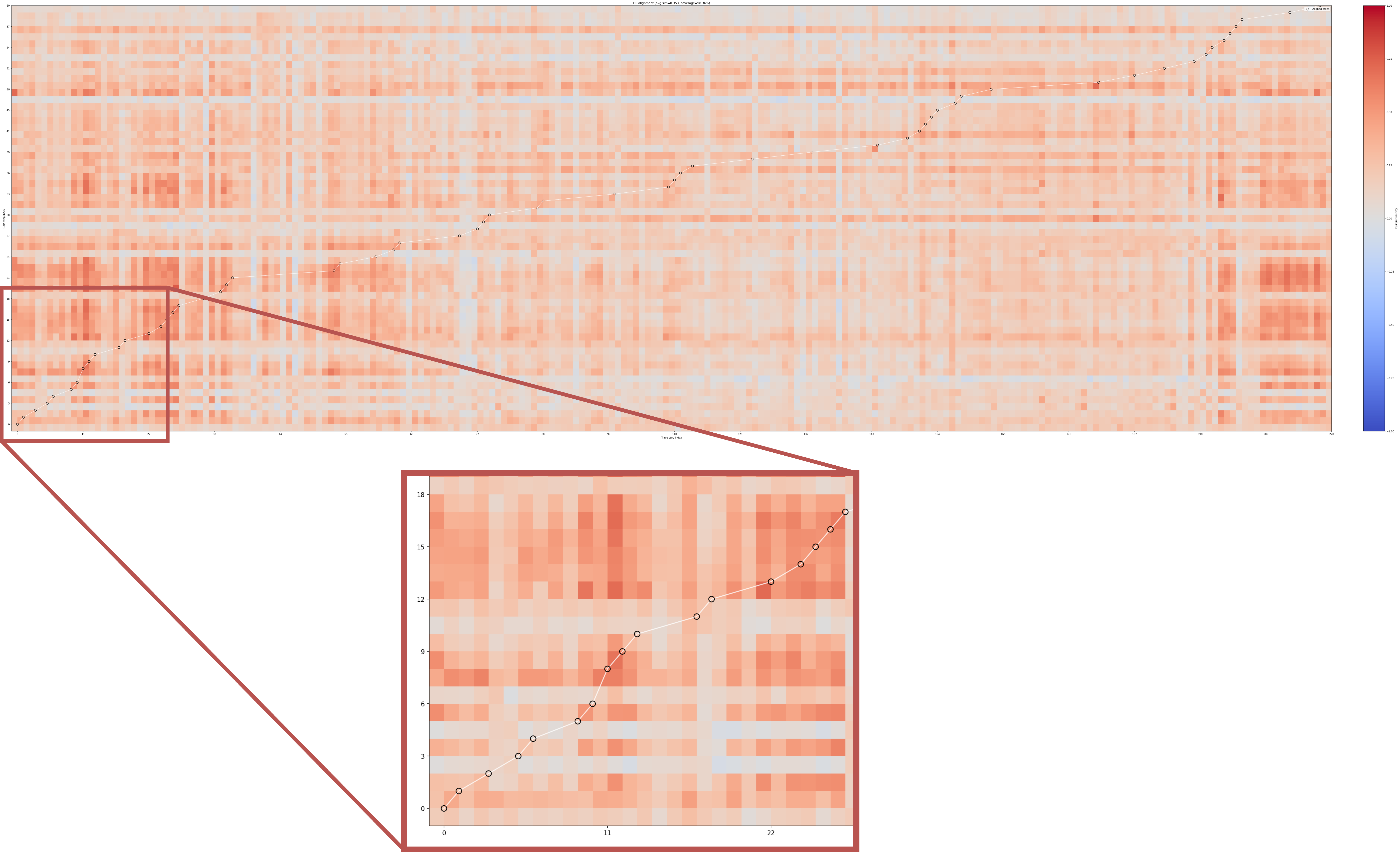}
  \caption{Visaulization of the Smith-Waterman Alignment Diagram for GPT-OSS 20B on AIME 2020 Question 1. Y-axis: golden step index, X-axis: trace step index, Heatmap: Cosine-Similarity.}
  \Description{}
\end{figure}

\section{Detailed Descriptions of Pseudo-code Function}
\label{app:codefunc}

\begin{algorithm}[H]
\footnotesize
\caption{\textsc{Segment}}
\begin{algorithmic}[1]
\Require text (string)
\Ensure list of step strings
\Function{Segment}{text}
  \If{\textit{text} is \textsc{None} or empty}
    \State \Return $[\,]$
  \EndIf
  \State $clean \gets$ \Call{StripInlineMathMarkers}{\textit{text}}
  \State \Comment{remove \texttt{\textbackslash(} \texttt{\textbackslash)} , \texttt{\textbackslash[} \texttt{\textbackslash]} , \texttt{\$}, all backslashes, and the literal string \texttt{<think>}}
  \State $segments \gets [\,]$
  \ForAll{$ln \in$ \Call{SplitLines}{$clean$}}
    \State $ln \gets$ \Call{Trim}{$ln$} \Comment{Trim = strip() in python}
    \If{$ln = \emptyset$}
      \State \textbf{continue}
    \EndIf
    \State $pieces \gets$ \Call{RegexSplit}{$ln$, \texttt{(?<=\.)\textbackslash s+}}
    \ForAll{$p \in pieces$}
      \State $p \gets$ \Call{Trim}{$p$}
      \If{$p \neq \emptyset$}
        \State \Call{Append}{$segments$, $p$}
      \EndIf
    \EndFor
  \EndFor
  \State \Return $segments$
\EndFunction
\end{algorithmic}
\end{algorithm}

\begin{algorithm}[H]
\footnotesize
\caption{\textsc{BuildGraph}}
\begin{algorithmic}[1]
\Require embeddings $X \in \mathbb{R}^{n\times d}$,\ cluster budget $k$ (default $200$)
\Ensure discrete path $\text{path}\in\{0,\dots,k'-1\}^n$; edge distances $\text{dist}\in\mathbb{R}_+^{n-1}$
\Function{BuildGraph}{$X,k$}
  \State $n \gets \mathrm{rows}(X)$; \ $k' \gets \min(k,n)$
  \State $(\text{labels},\text{centers}) \gets \Call{KMeans}{X,\ n\_clusters=k',\ \text{seed}=0}$
  \State $\text{path} \gets$ integer copy of $\text{labels}$
  \State $\text{dist} \gets [\,]$
  \For{$t \gets 1$ \textbf{to} $n-1$}
    \State $u \gets \text{path}[t-1]$; \ $v \gets \text{path}[t]$
    \State $d \gets \left\lVert \text{centers}[v] - \text{centers}[u] \right\rVert_2$
    \State append $d$ to $\text{dist}$
  \EndFor
  \State \Return $(\text{path},\ \text{dist})$
\EndFunction
\end{algorithmic}
\end{algorithm}

\begin{algorithm}[H]
\footnotesize
\caption{\textsc{Align}}
\begin{algorithmic}[1]
\Require model embeddings $X^{(r)} = (x^{(r)}_0,\dots,x^{(r)}_{n-1})$, gold embeddings $X^{(s)} = (x^{(s)}_0,\dots,x^{(s)}_{m-1})$, gap penalty $\lambda \ge 0$
\Ensure aligned index pairs $\mathcal{A} \subseteq \{0,\dots,n-1\}\times\{0,\dots,m-1\}$; average similarity $\mathrm{score}$; gold coverage $\mathrm{coverage}$
\Function{Align}{$X^{(r)}, X^{(s)}, \lambda$}
  \State $n \gets |X^{(r)}|,\quad m \gets |X^{(s)}|$
  \State \Comment{Cosine similarity helper: $\mathrm{CosSim}(a,b) = \frac{a}{\|a\|+\varepsilon}\cdot\frac{b}{\|b\|+\varepsilon}$}
  \State Construct similarity matrix $S \in \mathbb{R}^{n\times m}$ where $S[i,j] \gets \mathrm{CosSim}(x^{(r)}_i, x^{(s)}_j)$ for all $0\le i<n$, $0\le j<m$
  \State Initialize DP table $dp \in \mathbb{R}^{(n+1)\times(m+1)} \gets -\infty$ and backpointer table $bt \in \{0,\dots,n\}\times\{0,\dots,m\}^{(n+1)\times(m+1)}$
  \State $dp[0,0] \gets 0$
  \For{$i \gets 0$ to $n$}
    \For{$j \gets 0$ to $m$}
      \If{$i<n$ \textbf{and} $j<m$}
        \State \Comment{match (diagonal move)}
        \If{$dp[i,j] + S[i,j] > dp[i+1,j+1]$}
          \State $dp[i+1,j+1] \gets dp[i,j] + S[i,j]$; \quad $bt[i+1,j+1] \gets (i,j)$
        \EndIf
      \EndIf
      \If{$i<n$}
        \State \Comment{skip a model step (vertical move)}
        \If{$dp[i,j] - \lambda > dp[i+1,j]$}
          \State $dp[i+1,j] \gets dp[i,j] - \lambda$; \quad $bt[i+1,j] \gets (i,j)$
        \EndIf
      \EndIf
      \If{$j<m$}
        \State \Comment{skip a gold step (horizontal move)}
        \If{$dp[i,j] - \lambda > dp[i,j+1]$}
          \State $dp[i,j+1] \gets dp[i,j] - \lambda$; \quad $bt[i,j+1] \gets (i,j)$
        \EndIf
      \EndIf
    \EndFor
  \EndFor
  \State \Comment{Backtrack from $(n,m)$ to recover aligned pairs}
  \State $i \gets n,\ j \gets m,\ \mathcal{A} \gets [\,]$
  \While{$i>0$ \textbf{or} $j>0$}
    \State $(p_i,p_j) \gets bt[i,j]$
    \If{$p_i = i-1$ \textbf{and} $p_j = j-1$}
      \State append $(i-1,\ j-1)$ to $\mathcal{A}$
    \EndIf
    \State $i \gets p_i;\ j \gets p_j$
  \EndWhile
  \State reverse $\mathcal{A}$
  \State \Comment{Aggregate metrics}
  \State $\mathrm{score} \gets \begin{cases}
      \frac{1}{|\mathcal{A}|}\sum\limits_{(i,j)\in \mathcal{A}} S[i,j], & |\mathcal{A}|>0\\
      0, & \text{otherwise}
    \end{cases}$
  \State $\mathrm{coverage} \gets \dfrac{\left|\{\,j : (i,j)\in \mathcal{A}\,\}\right|}{\max(1,m)}$
  \State \Return $(\mathcal{A},\ \mathrm{score},\ \mathrm{coverage})$
\EndFunction
\end{algorithmic}
\end{algorithm}

\begin{algorithm}[H]
\footnotesize	
\caption{\textsc{AnalyzeGraph}}
\begin{algorithmic}[1]
\Require discrete path \texttt{path} (length $n$), edge distances \texttt{dist} (length $n-1$)
\Ensure \texttt{has\_loop} (bool), \texttt{loop\_count} (int), \texttt{diam} ($\mathbb{R}_{\ge 0}$), $\overline{C}$, $\overline{L}$, $\sigma$
\Function{AnalyzeGraph}{\texttt{path}, \texttt{dist}}
  \State \Comment{Build directed adjacency from consecutive labels}
  \State $\texttt{adj} \gets$ empty map $u \mapsto$ list of $(v,w)$
  \For{$(u,v,w)$ \textbf{in} \texttt{zip(path, path[1:], dist)}}
    \If{$u \neq v$}
      \State append $(v,w)$ to \texttt{adj[$u$]}
    \EndIf
  \EndFor
  \State \Comment{Detect first loop and count revisits of its entry node}
  \State $\texttt{seen}\gets\emptyset$; \texttt{has\_loop}$\gets$False; \texttt{loop\_count}$\gets 0$
  \For{$node$ \textbf{in} \texttt{path}}
    \If{$node \in \texttt{seen}$}
      \State \texttt{has\_loop}$\gets$True
      \State \texttt{loop\_count}$\gets$ occurrences of $node$ in \texttt{path} minus $1$
      \State \textbf{break}
    \EndIf
    \State add $node$ to \texttt{seen}
  \EndFor
  \State \Comment{Single-source shortest paths on the directed graph (Dijkstra)}
  \Function{Dijkstra}{$s$}
    \State $dist\_map \gets \{s\!\mapsto\!0\}$; priority queue $\mathcal{H}\gets[(0,s)]$
    \While{$\mathcal{H}$ not empty}
      \State $(d,u)\gets$ pop-min from $\mathcal{H}$
      \For{each $(v,w)$ in \texttt{adj[$u$]}}
        \If{$v\notin dist\_map$ \textbf{or} $d{+}w<dist\_map[v]$}
          \State $dist\_map[v]\gets d{+}w$; push $(dist\_map[v],v)$
        \EndIf
      \EndFor
    \EndWhile
    \State \Return $dist\_map$
  \EndFunction
  \State $D \gets [\,\textsc{Dijkstra}(u)\,:\,u\in\text{keys}(\texttt{adj})\,]$
  \State \texttt{diam} $\gets \max\{\max(\text{vals}(d)) : d\in D\}$ (default $0$ if empty)
  \State $\overline{L} \gets \dfrac{\sum_{d\in D}\sum\text{vals}(d)}{\sum_{d\in D}(\lvert d\rvert - 1)}$ \quad (if denominator $>0$, else $0$)
  \State \Comment{Undirected view for clustering coefficient}
  \State $\texttt{ud}\gets$ map $u\mapsto$ set of neighbors
  \For{each $u$ and $(v,{-})$ in \texttt{adj}}
    \State add $v$ to \texttt{ud[$u$]}; add $u$ to \texttt{ud[$v$]}
  \EndFor
  \State $sumC\gets 0$;\ $cnt\gets 0$
  \For{each $u$ with neighbors $N=\texttt{ud}[u]$}
    \If{$|N|<2$}
      \State \textbf{continue}
    \EndIf
    \State $E\_{\!N}\gets\#\{\{v,w\}\subset N: w\in \texttt{ud}[v]\}$
    \State $C(u)\gets \dfrac{E\_{\!N}}{\binom{|N|}{2}}$; \ $sumC{+}{=}\,C(u)$; \ $cnt{+}{=}1$
  \EndFor
  \State $\overline{C}\gets \dfrac{sumC}{cnt}$ \ (if $cnt>0$, else $0$)
  \State \Comment{Small-world index $\sigma=\dfrac{\overline{C}/C\_{\text{rand}}}{\overline{L}/L\_{\text{rand}}}$}
  \State $N\gets |\text{keys}(\texttt{ud})|$;\quad $K\gets \dfrac{\sum\_{u}|\,\texttt{ud}[u]\,|}{N}$ \ (if $N>0$, else $0$)
  \State $C\_{\text{rand}}\gets \dfrac{K}{N-1}$ \ (if $N>1$, else $0$)
  \State $L\_{\text{rand}}\gets \dfrac{\ln N}{\ln K}$ \ (if $N>1$ and $K>1$, else $\infty$)
  \State $C\_{\text{norm}}\gets \dfrac{\overline{C}}{C\_{\text{rand}}}$ \ (if $C\_{\text{rand}}>0$, else $0$)
  \State $L\_{\text{norm}}\gets \dfrac{\overline{L}}{L\_{\text{rand}}}$ \ (if $L\_{\text{rand}}<\infty$, else $0$)
  \State $\sigma \gets \dfrac{C\_{\text{norm}}}{L\_{\text{norm}}}$ \ (if $L\_{\text{norm}}>0$, else $0$)
  \State \Return (\texttt{has\_loop}, \texttt{loop\_count}, \texttt{diam}, $\overline{C}$, $\overline{L}$, $\sigma$)
\EndFunction
\end{algorithmic}
\end{algorithm}

\section{Variance Inflation Factor Table for Topological Features}

\begin{table}[H]
\scriptsize
\centering
\caption{TDA VIF by model.}
\label{tab:vif}
\resizebox{\linewidth}{!}{%
\begin{tabular}{lcccccccc}
\toprule
TDA Feature & Qwen3-8B & Qwen3-32B & Qwen3-235B & Deepseek-r1-7B & Deepseek-r1-32B & Deepseek-r1-70B & GPT-OSS-20B & GPT-OSS-120b \\
\midrule
H0 landscape max & inf & inf & inf & inf & inf & inf & inf & inf \\
H0 landscape mean & inf & inf & inf & inf & inf & inf & inf & inf \\
H0 max life & inf & inf & inf & inf & inf & inf & inf & inf \\
H1 landscape max & inf & inf & inf & inf & inf & inf & inf & inf \\
H1 max life & inf & inf & inf & inf & inf & inf & inf & inf \\
H0 landscape area & 760.3 & 563.6 & 1309.0 & 602.3 & 699.5 & 583.3 & 551.4 & 673.3 \\
H0 total life & 135.0 & 196.5 & 111.6 & 150.6 & 208.3 & 129.3 & 86.5 & 75.0 \\
H0 count & 106.5 & 173.6 & 115.0 & 132.5 & 183.9 & 134.8 & 103.0 & 193.6 \\
H0 entropy & 122.5 & 116.3 & 188.8 & 105.5 & 146.2 & 176.8 & 178.9 & 222.5 \\
H1 count & 168.8 & 141.3 & 74.1 & 105.0 & 134.5 & 117.6 & 141.6 & 142.1 \\
H1 total life & 150.0 & 171.6 & 41.3 & 76.9 & 123.9 & 59.3 & 114.5 & 99.9 \\
H0 mean life & 98.7 & 130.8 & 65.3 & 77.6 & 76.5 & 63.6 & 62.7 & 49.5 \\
H1 entropy & 63.2 & 30.7 & 69.7 & 37.5 & 68.2 & 76.9 & 68.3 & 84.0 \\
H0 betti centroid & 63.2 & 43.0 & 31.8 & 28.4 & 37.4 & 31.9 & 45.0 & 34.2 \\
H1 mean life & 33.9 & 42.7 & 13.4 & 29.3 & 26.0 & 16.8 & 17.7 & 7.5 \\
H0 betti width & 22.1 & 22.3 & 24.0 & 20.7 & 16.6 & 27.2 & 22.5 & 15.9 \\
H0 betti spread & 28.2 & 13.2 & 5.4 & 7.3 & 11.0 & 10.0 & 15.0 & 11.1 \\
H1 betti peak & 10.1 & 7.5 & 11.8 & 8.3 & 9.3 & 9.3 & 11.1 & 14.0 \\
H1 landscape area & 11.8 & 6.5 & 11.9 & 12.7 & 9.1 & 9.9 & 10.6 & 10.9 \\
H0 skewness & 18.8 & 12.2 & 10.5 & 7.2 & 8.0 & 9.0 & 8.3 & 6.1 \\
H1 max death & 6.1 & 6.0 & 4.7 & 6.0 & 6.0 & 5.1 & 7.4 & 6.6 \\
H1 skewness & 5.9 & 5.4 & 6.6 & 4.5 & 5.8 & 6.0 & 5.5 & 7.4 \\
H1 max birth & 5.7 & 5.9 & 4.6 & 5.3 & 5.7 & 4.4 & 7.1 & 6.4 \\
H1 landscape mean & 6.8 & 5.2 & 5.7 & 8.7 & 4.9 & 4.9 & 5.7 & 5.8 \\
H1 betti centroid & 4.9 & 5.5 & 3.8 & 5.6 & 4.5 & 4.3 & 7.0 & 5.2 \\
H1 betti location & 2.6 & 2.5 & 2.4 & 2.7 & 3.1 & 2.5 & 4.4 & 3.1 \\
H1 betti spread & 4.4 & 4.3 & 4.1 & 4.0 & 2.7 & 4.0 & 6.7 & 4.6 \\
H1 betti width & 3.5 & 3.2 & 3.2 & 2.7 & 2.4 & 2.8 & 3.0 & 2.1 \\
\bottomrule
\end{tabular}
}
\end{table}

\section{TDA Feature Clustering and Full Regression Details}
\label{app:clustered_tda}

To stabilize inference under multicollinearity, we cluster the 28 TDA features and regress Smith--Waterman alignment on cluster representatives.

\begin{figure}[H]
    \centering
    \includegraphics[width=0.8\linewidth]{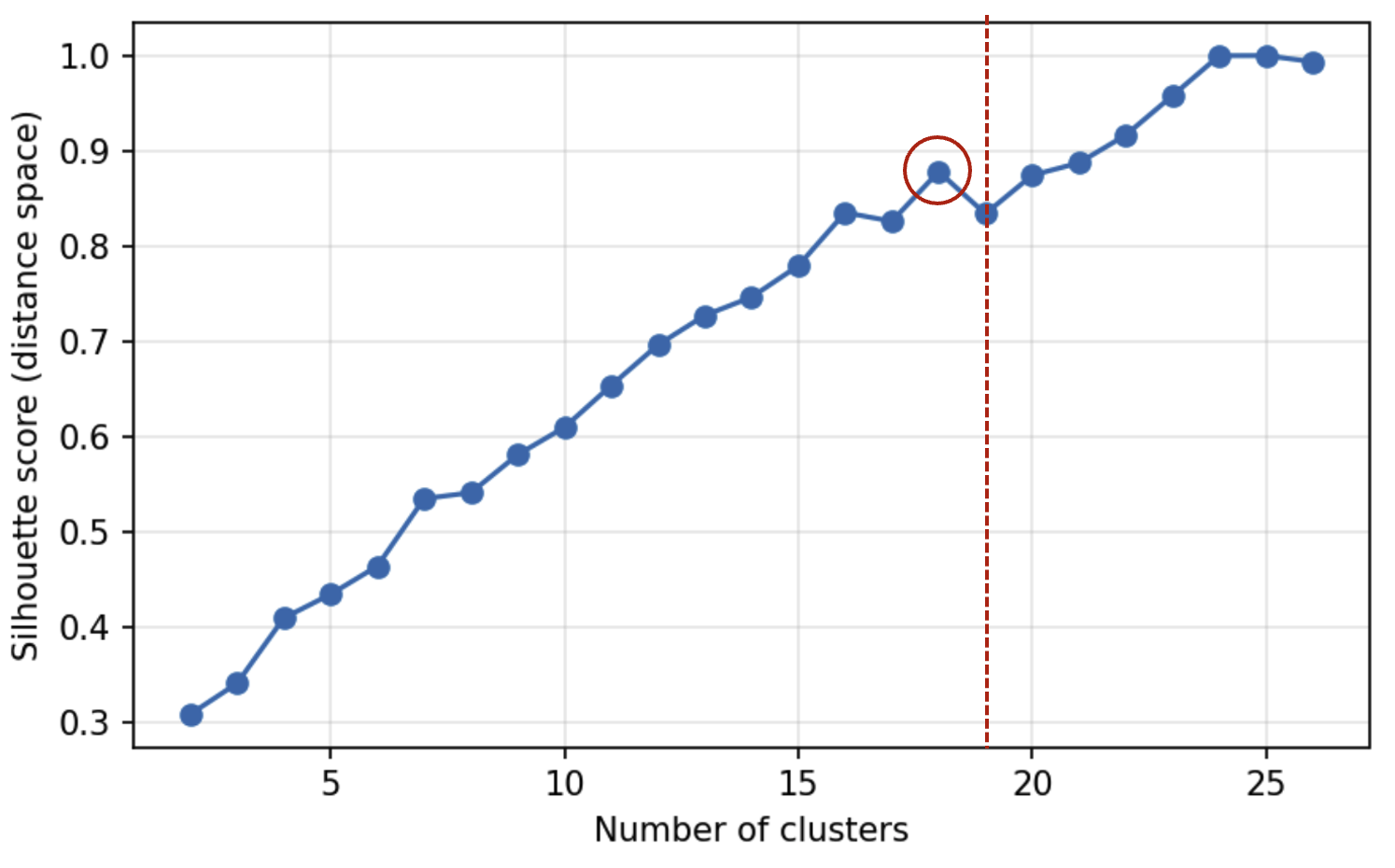}
    \caption{Silhouette score across candidate numbers of correlation-based feature clusters.}
    \label{fig:tda_cluster_silhouette}
    \Description{}
\end{figure}

\begin{table}[H]
\footnotesize
\centering
\caption{Feature membership for the selected 18 clusters.}
\label{tab:tda_clusters_multirow}
\begin{tabular}{cl}
\toprule
\textbf{Cluster} & \textbf{Feature} \\
\midrule
1 & H0 betti\_centroid \\
2 & H0 betti\_spread \\
3 & H0 betti\_width \\
4 & H0 count \\
4 & H0 entropy \\
4 & H0 total\_life \\
4 & H1 count \\
4 & H1 entropy \\
4 & H1 total\_life \\
5 & H0 landscape\_area \\
5 & H0 landscape\_max \\
5 & H0 landscape\_mean \\
5 & H0 max\_life \\
6 & H0 mean\_life \\
7 & H0 skewness \\
8 & H1 betti\_centroid \\
9 & H1 betti\_location \\
10 & H1 betti\_peak \\
11 & H1 betti\_spread \\
12 & H1 betti\_width \\
13 & H1 landscape\_area \\
14 & H1 landscape\_max \\
14 & H1 max\_life \\
15 & H1 landscape\_mean \\
16 & H1 max\_birth \\
16 & H1 max\_death \\
17 & H1 mean\_life \\
18 & H1 skewness \\
\bottomrule
\end{tabular}
\end{table}

\paragraph{Clustering procedure.}
Let $X\in\mathbb{R}^{N\times p}$ be the matrix of $p=28$ TDA features across $N=1440$ observations. We standardize each feature:
\[
\tilde x_{ij} = \frac{x_{ij}-\mu_j}{\sigma_j},\qquad
\mu_j=\frac{1}{N}\sum_{i=1}^N x_{ij},\qquad
\sigma_j=\sqrt{\frac{1}{N}\sum_{i=1}^N(x_{ij}-\mu_j)^2}.
\]
Using $R=\mathrm{Corr}(\tilde X)$, we define correlation distance $D=\mathbf{1}-|R|$ and run agglomerative clustering with average linkage. For clusters $A,B$,
\[
\Delta(A,B)=\frac{1}{|A|\,|B|}\sum_{i\in A}\sum_{j\in B}D_{ij}.
\]
At each $K$, we score partitions with the silhouette criterion and select $K^\star=18$ based on the stability--interpretability trade-off in Figure~\ref{fig:tda_cluster_silhouette}.

Cluster representatives are the averages of standardized members:
\[
g_r(i)=\frac{1}{|C_r|}\sum_{j\in C_r}\tilde x_{ij},\qquad r=1,\dots,18.
\]

\begin{table*}[h]
\scriptsize
\centering
\caption{Regression by Cluster (Across all models)}
\label{tab:tda_corrclusters_overall_horizontal}
\setlength{\tabcolsep}{3.2pt}
\begin{tabular}{lcccccccccccccccccc}
\toprule
& \textbf{C1} & \textbf{C2} & \textbf{C3} & \textbf{C4} & \textbf{C5} & \textbf{C6} & \textbf{C7} & \textbf{C8} & \textbf{C9} & \textbf{C10} & \textbf{C11} & \textbf{C12} & \textbf{C13} & \textbf{C14} & \textbf{C15} & \textbf{C16} & \textbf{C17} & \textbf{C18} \\
\midrule
Coef & 0.000 & \textbf{0.010**} & \textbf{-0.018**} & 0.004 & -0.001 & 0.002 & -0.006 & 0.001 & -0.004 & -0.001 & -0.001 & \textbf{0.007***} & -0.006 & 0.004 & -0.001 & \textbf{-0.004*} & 0.002 & -0.001 \\
(SE) & (0.009) & (0.005) & (0.007) & (0.006) & (0.005) & (0.010) & (0.005) & (0.004) & (0.003) & (0.005) & (0.003) & (0.003) & (0.005) & (0.005) & (0.004) & (0.003) & (0.004) & (0.004) \\
\bottomrule
\addlinespace
\multicolumn{10}{l}{Constant: 0.427*** (0.002), Observations: 1440 (180 $\times$ 8), $R^{2}$: 0.073, Adj.\ $R^{2}$: 0.061.} & \multicolumn{9}{r}{\textit{Notes:} Standard errors (SE) in parentheses. $^{*}p<0.10$, $^{**}p<0.05$, $^{***}p<0.01$.} \\
\end{tabular}
\end{table*}
\section{Features Regression}

\begin{table}[H]
\scriptsize
\centering
\caption{OLS Regression Results for TDA Features across all models}
\begingroup
\setlength{\tabcolsep}{3pt} 
\renewcommand{\arraystretch}{1.1}

\resizebox{\linewidth}{!}{%
\begin{tabular}{l*{10}{c}}
\hline
 & const & H0 betti centroid & H0 betti spread & H0 betti width & H0 count & H0 entropy & H0 landscape area & H0 landscape max & H0 landscape mean & H0 max life \\
\hline
Coef & 0.591 & -0.404 & \textbf{0.838$^{**}$} & \textbf{-0.376$^{***}$} & \textbf{0.001$^{*}$} & -0.020 & -1.158 & 0.108 & 0.036 & 0.215 \\
(SE)  & (0.472) & (0.424) & (0.382) & (0.139) & (0.000) & (0.066) & (2.534) & (0.363) & (0.121) & (0.725) \\
\hline
\end{tabular}
}

\vspace{0.5em}

\resizebox{\linewidth}{!}{%
\begin{tabular}{l*{10}{c}}
\hline
  & H0 mean life & H0 skewness & H0 total life & H1 betti centroid & H1 betti location & H1 betti peak & H1 betti spread & H1 betti width & H1 count & H1 entropy \\
\hline
Coef & \textbf{0.755$^{*}$} & -0.019 & -0.002 & 0.007 & -0.031 & -0.001 & -0.063 & \textbf{0.030$^{**}$} & -0.000 & -0.016 \\
(SE) & (0.405) & (0.015) & (0.001) & (0.052) & (0.021) & (0.001) & (0.105) & (0.015) & (0.000) & (0.029) \\
\hline
\end{tabular}
}

\vspace{0.5em}

\resizebox{\linewidth}{!}{%
\begin{tabular}{l*{9}{c}}
\hline
 & H1 landscape area & H1 landscape max & H1 landscape mean & H1 max birth & H1 max death & H1 max life & H1 mean life & H1 skewness & H1 total life \\
\hline
Coef & -0.941 & 0.046 & -0.138 & \textbf{0.205$^{*}$} & \textbf{-0.366$^{***}$} & 0.092 & -1.253 & -0.005 & 0.014 \\
(SE)  & (1.196) & (0.064) & (0.608) & (0.106) & (0.122) & (0.129) & (1.146) & (0.012) & (0.009) \\
\hline
\end{tabular}
}

\vspace{0.5em}

\begin{tabular}{lcccc}
\textit{Observations:} 1440 &
\textit{$R^2$:} 0.085 &
\textit{Adjusted $R^2$:} 0.069 &
\textit{Residual Std. Error:} 0.066 (df=1414) &
\textit{F Statistic:} 5.264$^{***}$ (df=25; 1414) \\
\end{tabular}

\vspace{0.25em}

\begin{tabular}{l}
\textit{Note:} $^{*}$p$<$0.1; $^{**}$p$<$0.05; $^{***}$p$<$0.01 \\
\end{tabular}
\endgroup
\end{table}

\begin{table}[H] 
\scriptsize
\centering
\caption{OLS Regression Results for Graph Features across all models}
\begin{tabular}{lccccccc}
\hline
 & Const & Avg clustering & Avg path length & Diameter & Has loop & Loop count & Small world index \\
\hline
Coef & 0.481$^{***}$ & -0.327 & 0.000 & \textbf{-0.000$^{*}$} & \textbf{-0.022$^{**}$} & -0.000 & 0.019 \\
(SE)        & (0.013)        & (0.382) & (0.000)          & (0.000)        & (0.011)          & (0.001)           & (0.018) \\
\hline
Observations         & \multicolumn{7}{c}{1440} \\
$R^2$                & \multicolumn{7}{c}{0.034} \\
Adjusted $R^2$       & \multicolumn{7}{c}{0.030} \\
Residual Std. Error  & \multicolumn{7}{c}{0.068 (df=1433)} \\
F Statistic          & \multicolumn{7}{c}{8.295$^{***}$ (df=6; 1433)} \\
\hline
\multicolumn{8}{r}{\textit{Note:} $^{*}$p$<$0.1; $^{**}$p$<$0.05; $^{***}$p$<$0.01} \\
\end{tabular}
\end{table}

\newpage
\section{Regression Results for TDA Features vs Graph Features}

\begin{table}[H]
\centering
\scriptsize
\caption{Regression results for TDA Features vs Graph Features over all models (1440 Observations)}
\label{tab:tdavsgraph}
\begin{tabular}{lccccc}
\toprule
 & Graph avg. clustering & Graph avg. path length & Graph diameter & Graph loop count & Graph small world index \\
\midrule
H0 betti centroid & \colorbox{orange}{-0.209***} & \colorbox{orange}{273.643***} & \colorbox{orange}{970.601***} & \colorbox{yellow}{25.866**} & \colorbox{orange}{-5.115***} \\
 & (0.080) & (78.988) & (227.813) & (12.244) & (1.702) \\
H0 betti spread & 0.054 & \colorbox{orange}{-321.296***} & \colorbox{orange}{-899.377***} & 2.465 & 0.574 \\
 & (0.072) & (71.034) & (204.872) & (11.011) & (1.531) \\
H0 betti width & \colorbox{lime}{-0.051*} & 14.468 & 67.202 & 5.293 & \colorbox{yellow}{-1.159**} \\
 & (0.026) & (25.795) & (74.398) & (3.998) & (0.556) \\
H0 count & \colorbox{orange}{0.001***} & \colorbox{orange}{-0.814***} & \colorbox{orange}{-2.337***} & -0.010 & \colorbox{orange}{0.019***} \\
 & (0.000) & (0.065) & (0.187) & (0.010) & (0.001) \\
H0 entropy & \colorbox{orange}{-0.063***} & \colorbox{orange}{88.214***} & \colorbox{orange}{276.084***} & -1.671 & \colorbox{orange}{-1.745***} \\
 & (0.012) & (12.261) & (35.364) & (1.901) & (0.264) \\
H0 landscape area & \colorbox{yellow}{-1.056**} & \colorbox{lime}{838.202*} & \colorbox{yellow}{2685.715**} & -11.358 & \colorbox{yellow}{-20.324**} \\
 & (0.478) & (471.764) & (1360.639) & (73.127) & (10.167) \\
H0 landscape max & \colorbox{lime}{0.122*} & \colorbox{lime}{-111.890*} & \colorbox{lime}{-333.755*} & 7.052 & 2.149 \\
 & (0.068) & (67.502) & (194.685) & (10.463) & (1.455) \\
H0 landscape mean & \colorbox{lime}{0.041*} & \colorbox{lime}{-37.297*} & \colorbox{lime}{-111.252*} & 2.351 & 0.716 \\
 & (0.023) & (22.501) & (64.895) & (3.488) & (0.485) \\
H0 max life & \colorbox{lime}{0.245*} & \colorbox{lime}{-223.779*} & \colorbox{lime}{-667.510*} & 14.103 & 4.297 \\
 & (0.137) & (135.003) & (389.370) & (20.927) & (2.910) \\
H0 mean life & \colorbox{orange}{0.344***} & \colorbox{orange}{-263.123***} & \colorbox{orange}{-865.344***} & \colorbox{orange}{-40.282***} & \colorbox{orange}{9.313***} \\
 & (0.076) & (75.437) & (217.572) & (11.693) & (1.626) \\
H0 skewness & -0.002 & \colorbox{orange}{13.296***} & \colorbox{orange}{35.447***} & \colorbox{yellow}{-0.854*} & 0.009 \\
 & (0.003) & (2.839) & (8.189) & (0.440) & (0.061) \\
H0 total life & \colorbox{orange}{-0.001***} & \colorbox{orange}{0.993***} & \colorbox{orange}{2.465***} & 0.013 & \colorbox{orange}{-0.019***} \\
 & (0.000) & (0.251) & (0.724) & (0.039) & (0.005) \\
H1 betti centroid & \colorbox{lime}{0.019*} & \colorbox{lime}{15.890*} & 39.964 & 0.297 & 0.329 \\
 & (0.010) & (9.620) & (27.746) & (1.491) & (0.207) \\
H1 betti location & \colorbox{yellow}{-0.009**} & -3.503 & -4.385 & -0.141 & \colorbox{yellow}{-0.201**} \\
 & (0.004) & (3.896) & (11.236) & (0.604) & (0.084) \\
H1 betti peak & 0.000 & \colorbox{lime}{-0.201*} & \colorbox{yellow}{-0.675**} & 0.019 & 0.002 \\
 & (0.000) & (0.108) & (0.311) & (0.017) & (0.002) \\
H1 betti spread & \colorbox{lime}{-0.036*} & \colorbox{lime}{-37.993*} & \colorbox{yellow}{-114.178**} & 1.729 & -0.592 \\
 & (0.020) & (19.505) & (56.255) & (3.023) & (0.420) \\
H1 betti width & 0.003 & -2.905 & -10.460 & 0.433 & 0.075 \\
 & (0.003) & (2.800) & (8.076) & (0.434) & (0.060) \\
H1 count & \colorbox{lime}{-0.000*} & \colorbox{orange}{0.344***} & \colorbox{orange}{1.042***} & -0.003 & \colorbox{orange}{-0.007***} \\
 & (0.000) & (0.075) & (0.215) & (0.012) & (0.002) \\
H1 entropy & 0.000 & \colorbox{yellow}{-12.062**} & \colorbox{yellow}{-37.526**} & 1.197 & \colorbox{lime}{0.216*} \\
 & (0.005) & (5.355) & (15.444) & (0.830) & (0.115) \\
H1 landscape area & 0.034 & -61.925 & -227.329 & \colorbox{lime}{-65.896*} & 4.241 \\
 & (0.226) & (222.583) & (641.964) & (34.502) & (4.797) \\
H1 landscape max & -0.015 & -8.426 & -12.834 & 1.668 & -0.195 \\
 & (0.012) & (11.960) & (34.495) & (1.854) & (0.258) \\
H1 landscape mean & 0.006 & -15.865 & -183.841 & \colorbox{orange}{66.944***} & -1.628 \\
 & (0.115) & (113.188) & (326.451) & (17.545) & (2.439) \\
H1 max birth & 0.012 & -29.773 & -63.113 & 3.418 & 0.152 \\
 & (0.020) & (19.689) & (56.786) & (3.052) & (0.424) \\
H1 max death & -0.008 & \colorbox{yellow}{57.910**} & \colorbox{yellow}{147.447**} & 0.585 & -0.183 \\
 & (0.023) & (22.737) & (65.576) & (3.524) & (0.490) \\
H1 max life & -0.029 & -16.853 & -25.668 & 3.336 & -0.389 \\
 & (0.024) & (23.921) & (68.991) & (3.708) & (0.516) \\
H1 mean life & \colorbox{lime}{-0.388*} & \colorbox{yellow}{532.831**} & \colorbox{yellow}{1550.869**} & -27.164 & \colorbox{yellow}{-10.745**} \\
 & (0.216) & (213.265) & (615.088) & (33.058) & (4.596) \\
H1 skewness & \colorbox{lime}{0.004*} & 0.141 & -0.876 & 0.487 & 0.066 \\
 & (0.002) & (2.185) & (6.301) & (0.339) & (0.047) \\
H1 total life & \colorbox{orange}{0.005***} & \colorbox{orange}{-5.358***} & \colorbox{orange}{-15.633***} & 0.054 & \colorbox{orange}{0.114***} \\
 & (0.002) & (1.640) & (4.731) & (0.254) & (0.035) \\
const & 0.123 & -150.129* & -556.748** & -7.515 & 3.768** \\
 & (0.089) & (87.948) & (253.655) & (13.633) & (1.895) \\
\midrule
$R^2$ & 0.347 & 0.384 & 0.376 & 0.074 & 0.384 \\
Adjusted $R^2$ & 0.335 & 0.373 & 0.365 & 0.058 & 0.373 \\
\bottomrule
\addlinespace
\multicolumn{6}{l}
\textit{Notes:} Standard errors in parentheses. \colorbox{lime}{$^{*}p<0.10$}, \colorbox{yellow}{$^{**}p<0.05$}, \colorbox{orange}{$^{***}p<0.01$}.
\end{tabular}
\end{table}
\end{document}